\def\year{2020}\relax
\documentclass[letterpaper]{article} % DO NOT CHANGE THIS

\usepackage{aaai20}  % DO NOT CHANGE THIS
\usepackage{times}  % DO NOT CHANGE THIS
\usepackage{helvet} % DO NOT CHANGE THIS
\usepackage{courier}  % DO NOT CHANGE THIS
\usepackage[hyphens]{url}  % DO NOT CHANGE THIS
\usepackage{graphicx} % DO NOT CHANGE THIS
\usepackage{graphicx}  % DO NOT CHANGE THIS
\usepackage{amsmath}
\usepackage{floatrow}
\usepackage{multirow}
\usepackage{adjustbox}
\usepackage{subcaption}
\usepackage{ctable}
\usepackage{tabularx}
\usepackage{makecell}
\usepackage[ruled, noend]{algorithm2e}
\usepackage[scaled=.85]{beramono}
\usepackage{csquotes}
\usepackage{threeparttable}

\title{Spiking-YOLO: Spiking Neural Network for Energy-Efficient Object Detection}

\author{
Seijoon Kim, Seongsik Park, Byunggook Na, Sungroh Yoon\\
Department of Electrical and Computer Engineering, ASRI, INMC, and Institute of Engineering Research\\
Seoul National University, Seoul 08826, Korea\\
sryoon@snu.ac.kr}

\begin{document}

\maketitle

\begin{abstract}
Over the past decade, deep neural networks (DNNs) have demonstrated remarkable performance in a variety of applications. As we try to solve more advanced problems, increasing demands for computing and power resources has become inevitable. Spiking neural networks (SNNs) have attracted widespread interest as the third-generation of neural networks due to their event-driven and low-powered nature. SNNs, however, are difficult to train, mainly owing to their complex dynamics of neurons and non-differentiable spike operations. Furthermore, their applications have been limited to relatively simple tasks such as image classification. In this study, we investigate the performance degradation of SNNs in a more challenging regression problem (i.e., object detection). Through our in-depth analysis, we introduce two novel methods: channel-wise normalization and signed neuron with imbalanced threshold, both of which provide fast and accurate information transmission for deep SNNs. Consequently, we present a first spiked-based object detection model, called Spiking-YOLO. Our experiments show that Spiking-YOLO achieves remarkable results that are comparable (up to 98\%) to those of Tiny YOLO on non-trivial datasets, PASCAL VOC and MS COCO. Furthermore, Spiking-YOLO on a neuromorphic chip consumes approximately 280 times less energy than Tiny YOLO and converges 2.3 to 4 times faster than previous SNN conversion methods.
\end{abstract}

\section{Introduction}
One of the primary reasons behind the recent success of deep neural networks (DNNs) can be attributed to the development of high-performance computing systems and the availability of large amounts of data for model training. However, solving more intriguing and advanced problems in real-world applications requires more sophisticated models and training data, which results in significant increase in computational overhead and power consumption. To overcome these challenges, many researchers have attempted to design computationally- and energy-efficient DNNs using pruning \cite{guo2016dynamic,he2017channel}, compression \cite{han2015deep,kim2015compression}, and quantization \cite{gong2014compressing,park2018quantized}, some of which have shown promising results. Despite these efforts, the demand for computing and power resources will increase as deeper and more complicated neural networks achieve higher accuracy \cite{tan2019efficientnet}.

Spiking neural networks (SNNs), which are the third-generation neural networks, were introduced to mimic how information is encoded and processed in the human brain by employing spiking neurons as computation units \cite{maass1997networks}. Unlike conventional neural networks, SNNs transmit information via the precise timing (temporal) of spike trains consisting of a series of spikes (discrete), rather than a real value (continuous). That is, SNNs utilize temporal aspects in information transmission as in biological neural systems \cite{mainen1995reliability}, thus providing sparse yet powerful computing ability \cite{mostafa2017fast,bellec2018long}. Moreover, the spiking neurons integrate inputs into a membrane potential when spikes are received and generate (fire) spikes when the membrane potential reaches a certain threshold, which enables event-driven computation. Driven by the sparse nature of spike events and event-driven computation, SNNs offer exceptional power efficiency and are the preferred neural networks in neuromorphic architectures \cite{merolla2014million,poon2011neuromorphic}.

Despite their excellent potential, SNNs have been limited to relatively simple tasks (e.g., image classification) and small datasets (e.g., MNIST and CIFAR), on a rather shallow structure \cite{lee2016training,wu2019direct}. One of the primary reasons for the limited application scope is the lack of scalable training algorithms due to complex dynamics and non-differentiable operations of spiking neurons. DNN-to-SNN conversion methods, as an alternative approach, have been studied widely in recent years \cite{cao2015spiking,diehl2015fast,sengupta2019going}. These methods are based on the idea of importing pre-trained parameters (e.g., weights and biases) from a DNN to an SNN. DNN-to-SNN conversion methods have achieved comparable results in deep SNNs to those of original DNNs (e.g., VGG and ResNet); however, results from MNIST and CIFAR datasets were competitive, while those of ImageNet dataset were unsatisfactory when compared with DNN's accuracy.

In this study, we investigate a more advanced machine learning problem in deep SNNs, namely object detection, using DNN-to-SNN conversion methods. Object detection is regarded as significantly more challenging as it involves both recognizing multiple overlapped objects and calculating precise coordinates for bounding boxes. Thus, it requires high numerical precision in predicting the output values of neural networks (i.e., regression problem) instead of selecting one class with the highest probability (i.e., argmax function) as performed in image classification. Based on our in-depth analysis, several issues arise when object detection is applied in deep SNNs: a) inefficiency of conventional normalization methods and b) absence of an efficient implementation method of leaky-ReLU in an SNN domain. 

To overcome these issues, we introduce two novel methods; channel-wise normalization and signed neuron with imbalanced threshold. Consequently, we present a spike-based object detection model, called Spiking-YOLO. As the first step towards object detection in SNNs, we implemented Spiking-YOLO based on Tiny YOLO \cite{redmon2016you}. To the best of our knowledge, this is the first deep SNN for object detection that achieves comparable results to those of DNNs on non-trivial datasets, PASCAL VOC and MS COCO.
Our contributions can be summarized as follows:

\begin{itemize}
\item \textbf{First object detection model in deep SNNs} We present Spiking-YOLO, a model that enables energy-efficient object detection in deep SNNs, for the first time. Spiking-YOLO achieves comparable results to original DNNs on non-trivial datasets, i.e., 98\%.

\item \textbf{Channel-wise normalization} We developed a fine-grained normalization method for deep SNNs. The proposed method enables a higher, yet proper firing rate in multiple neurons, thus leads to fast and accurate information transmission in deep SNNs.
\item \textbf{Signed neuron featuring imbalanced threshold} We proposed an accurate and efficient implementation method of leaky-ReLU in an SNN domain. The proposed method can easily be implemented in neuromorphic chips with minimum overheads. 

\end{itemize}

\section{Related work}
\subsection{DNN-to-SNN conversion}
In contrary to DNNs, SNNs use spike trains consisting of a series of spikes to convey information between neurons.
The integrate-and-fire neurons accumulate input $z$ into a membrane potential $V_\textrm{mem}$ as
\begin{equation}
\label{eq:vmem}
    V_{\textrm{mem},j}^{l}(t) = V_{\textrm{mem},j}^{l}(t-1) + z_{j}^{l}(t) - V_{\textrm{th}}\Theta_{j}^{l}(t)\textrm{,}
\end{equation}
where $\Theta_{j}^{l}(t)$ is a spike, and $z_{j}^{l}(t)$ is the input of $j$th neuron in the $l$th layer with a threshold voltage $V_\textrm{th}$. $z_{j}^{l}(t)$ can be described as
\begin{equation}
\label{eq:psp}
z_{j}^{l}(t) = \sum_{i}{w_{i,j}^{l}\Theta_{i}^{l\textrm{-}1}(t)+b_{j}^{l}} \textrm{,}
\end{equation}
where $w$ and $b$ are weight and bias, respectively.
A spike $\Theta$ is generated when the integrated value $V_\textrm{mem}$ exceeds the threshold voltage $V_\textrm{th}$ as
\begin{equation}
\label{eq:spiking_func}
\Theta_{i}^{l}(t) = U(V_{\textrm{mem},i}^{l}(t) - V_{\textrm{th}}) \textrm{,}
\end{equation}
where $U(x)$ is a unit step function.
Due to the event-driven nature, SNNs offer energy-efficient operations \cite{pfeiffer2018deep}. However, they are difficult to train which has been one of the major obstacles when deploying SNNs in various applications \cite{wu2019direct}.

The training method of SNNs consists of unsupervised learning with spike-timing-dependent plasticity (STDP) \cite{diehl2015unsupervised} and supervised learning with gradient descent and error back-propagation \cite{lee2016training}. Although STDP is biologically more plausible, the learning performance is significantly lower than that of supervised learning. Recent works proposed a supervised learning algorithm with a function that approximates the non-differentiable portion (integrate-and-fire) of SNNs \cite{jin2018hybrid,lee2016training} to improve the learning performance. Despite these efforts, most previous works have been limited to the image classification task and MNIST dataset on shallow SNNs.

As an alternative approach, the conversion of DNNs to SNNs has been recently proposed. \cite{cao2015spiking} proposed a DNN-to-SNN conversion method that neglected bias and max-pooling. In subsequent work, \cite{diehl2015fast} proposed data-based normalization to improve the performance in deep SNNs. \cite{rueckauer2017conversion} presented an implementation method of batch normalization and spike max-pooling. \cite{sengupta2019going} expanded conversion methods to VGG and residual architectures. Nonetheless, most previous works have been limited to the image classification task \cite{park2018fast}. 

\subsection{Object detection}
Object detection locates a single or multiple object(s) in an image or video by drawing bounding boxes, then identifying their classes.
Hence, an object detection model consists of not only a classifier that classifies objects but also a regressor that predicts the precise coordinates (x- and y-axis) and size (width and height) of the bounding boxes. Because predicting precise coordinates of bounding boxes is critical, object detection is considered to be a much more challenging task than image classification, where an argmax function is used to simply pick one class with the highest probability.

Region-based CNN (R-CNN) \cite{girshick2014rich} is considered to be one of the most significant advances in object detection. To improve detection performance and speed, various extended versions of R-CNN have been proposed, namely fast R-CNN \cite{girshick2015fast}, faster R-CNN \cite{ren2015faster}, and Mask R-CNN \cite{he2017mask}. Nevertheless, R-CNN based networks suffer from a slow inference speed due to multiple-stage detection schemes and thus are not suitable for real-time object detection. 

As an alternative approach, one-stage detection methods have been proposed, where the bounding box information is extracted, and the objects are classified in a unified network. In one-stage detection models, \enquote{Single-shot multi-box detector} (SSD) \cite{liu2016ssd} and \enquote{You only look once} (YOLO)~\cite{redmon2018yolov3} achieve the state-of-the-art performance. Particularly, the YOLO has superior inference speed (FPS) without a significant loss of accuracy, which is a critical factor in real-time object detection. Thus we selected Tiny YOLO as our object detection model. 

\section{Methods}
In object detection, recognizing multiple objects and drawing bounding boxes around them (i.e., regression problem) poses great challenges: high numerical precision is required to predict the output value of the network. When object detection is applied in deep SNNs using conventional DNN-to-SNN conversion methods, it suffers from severe performance degradation and is unable to detect any objects. Our in-depth analysis highlights possible explanations for this performance degradation: a) an extremely low firing rate in numerous neurons and b) lack of an efficient implementation method of leaky-ReLU in SNNs. To overcome these complications, we propose two novel methods: channel-wise normalization and signed neuron with imbalanced threshold.

\subsection{Channel-wise data-based normalization}
\begin{figure}[t]
    \center
    \includegraphics[width=.96\columnwidth]{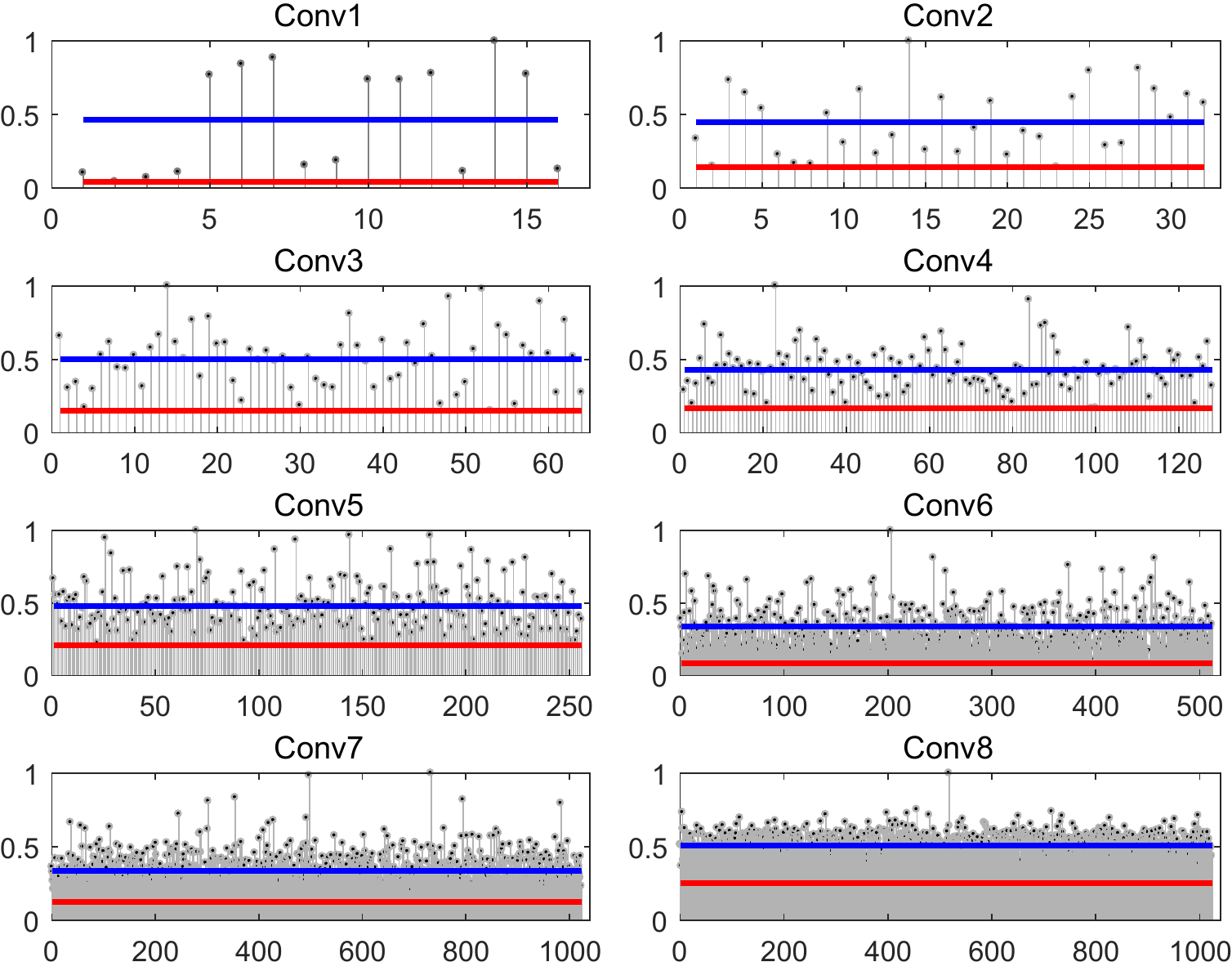}
	\caption{Normalized maximum activation via layer-wise normalization in each channel for eight convolutional layers in Tiny YOLO. Blue and red lines indicate the average and minimum of the normalized activations, respectively.}
	\label{fig:activation_distribution}
    %\vspace{-0.5em}
\end{figure}

\textbf{Conventional normalization methods } In a typical SNN, it is vital to ensure that a neuron generates spike trains according to the magnitude of the input and transmits those spike trains without any information loss. However, information loss can occur from under- or over-activation in the neurons given a fixed number of time steps. For instance, if a threshold voltage $V_\mathrm{th}$ is extremely large or the input is small, then a membrane potential $V_\mathrm{mem}$ will require a long time to reach $V_\mathrm{th}$, thus resulting in a low firing rate (i.e., under-activation). Conversely, if $V_\mathrm{th}$ is extremely small or input is large, then $V_\mathrm{mem}$ will most likely exceed $V_\mathrm{th}$ and the neuron will generate spikes regardless of the input value (i.e., over-activation). It is noteworthy that the firing rate can be defined as $\frac{N}{T}$, where $N$ is the total number of spikes in a given time step $T$. The maximum firing rate will be 100\% since a spike can be generated at every time step.

To prevent under- or over-activation in the neurons, both the weights and the threshold voltage need to be carefully chosen for sufficient and balanced activation of the neuron \cite{diehl2015fast}. Various data-based normalization methods \cite{diehl2015fast} have been proposed. Layer-wise normalization \cite{diehl2015fast} (abbreviated to layer-norm) is one of the most well-known normalization methods; layer-norm normalizes weights in a specific layer using the maximum activation of the corresponding layer, calculated from running the training dataset in a DNN. This is based on an assumption that the distributions of the training and test datasets are similar. In addition, note that normalizing the weights using the maximum activation will have the same effect as normalizing the output activation. Layer-norm can be calculated by

\begin{equation}
\label{eq:data-based norm}
\tilde{w}^{l} = w^{l} \frac{\lambda^{l-1}}{\lambda^{l}} \quad \textrm{and} \quad \tilde{b}^{l} = \frac{b^{l}}{\lambda^{l}},
\end{equation} where $w$, $\lambda$, and $b$ are the weights, the maximum activations calculated from the training dataset, and bias in layer $l$, respectively. As an extended version of layer-norm, \cite{rueckauer2017conversion} introduced an approach that normalizes the activations using the 99.9$\mathrm{th}$ percentile of the maximum activation; this increases the robustness to outliers and ensures sufficient firing of neurons. However, our experiments show that when object detection is applied in deep SNNs using the conventional normalization methods, the model suffers from significant performance degradation.

\begin{figure}[t]
    \center
    \includegraphics[width=0.95\columnwidth]{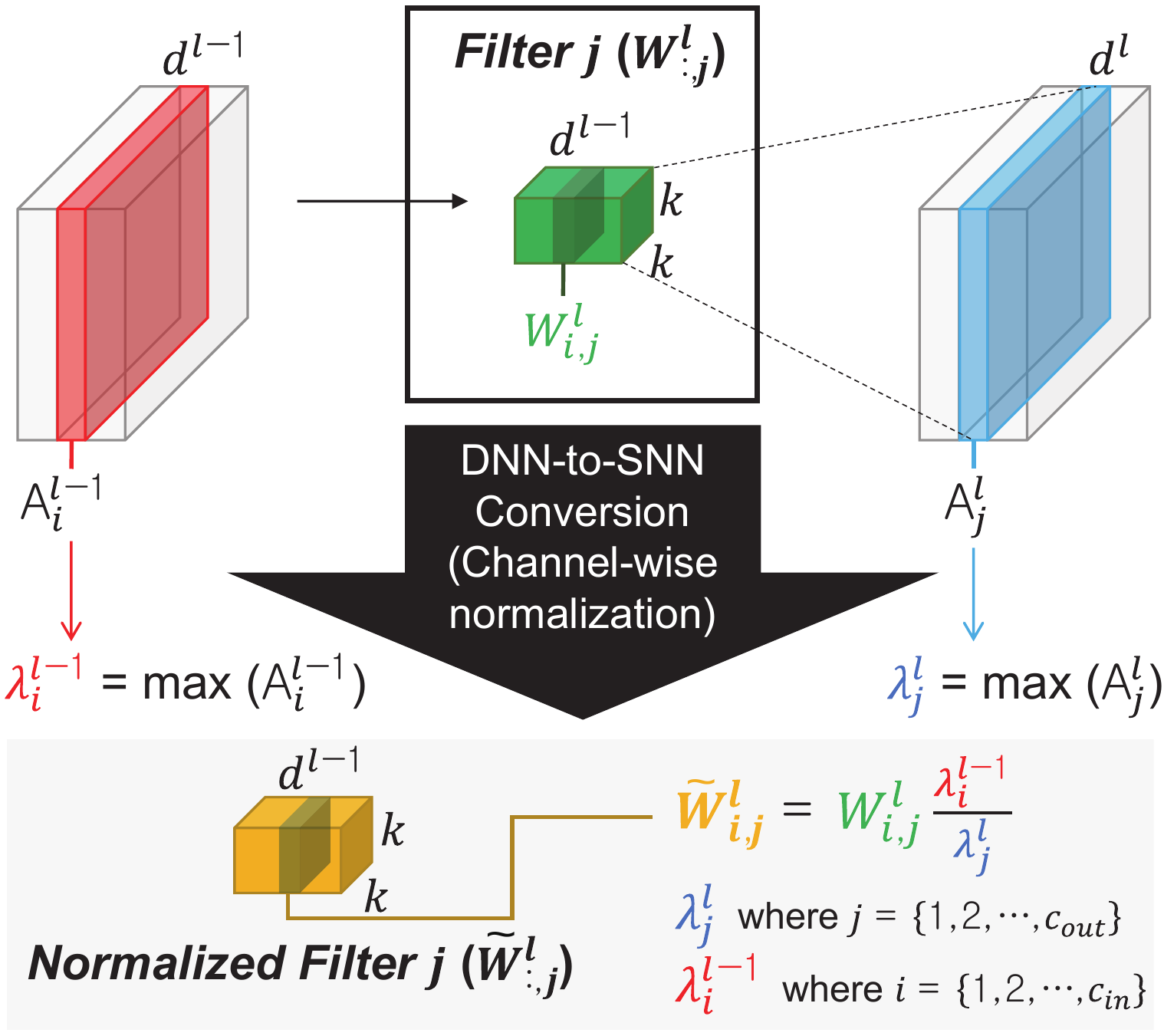}
    \caption{Proposed channel-wise normalization; $A^{l}_j$ is $j$th activation matrix (i.e., feature map) in layer $l$.}
	\label{fig:channel-norm}
    %\vspace{-0.5em}
\end{figure}

\textbf{Analysis of layer-norm limitation } Figure~\ref{fig:activation_distribution} represents the normalized maximum activation values in each channel obtained from layer-norm. Tiny YOLO consists of eight convolutional layers; the x-axis indicates the channel index and the y-axis represents the normalized maximum activation values. The blue and red lines indicate the average and minimum values of the normalized activations in each layer, respectively. As highlighted in Figure~\ref{fig:activation_distribution}, for a specific convolutional layer, the deviation of the normalized activations on each channel is relatively large. For example, in the Conv1 layer, the normalized maximum activation is close to 1 for certain channels (e.g., channels 6, 7, and 14) and 0 for other channels (e.g., channels 1, 2, 3, 13, and 16). The same can be said for the other convolutional layers. Clearly, layer-norm yields exceptionally small normalized activations (i.e., under-activation) in numerous channels that had relatively small activation values prior to the normalization. 

These extremely small normalized activations were undetected in image classification, but can be extremely problematic in solving regression problems in deep SNNs. For instance, to transmit 0.7, 7 spikes and 10 time steps are required. Applying the same logic, transmitting 0.007 would require 7 spikes and 1000 time steps without any loss of information. Hence, to send either extremely small (e.g., 0.007) or precise (e.g., 0.9007 vs. 0.9000) values without any loss, a large number of time steps is required. The number of time steps is considered as the resolution of the information being transmitted. Consequently, extremely small normalized activations yield low firing rates which results in information loss when the number of time steps is less than what it needs to be.

\begin{figure}[t]
    \centering
    \includegraphics[width=.98\columnwidth]{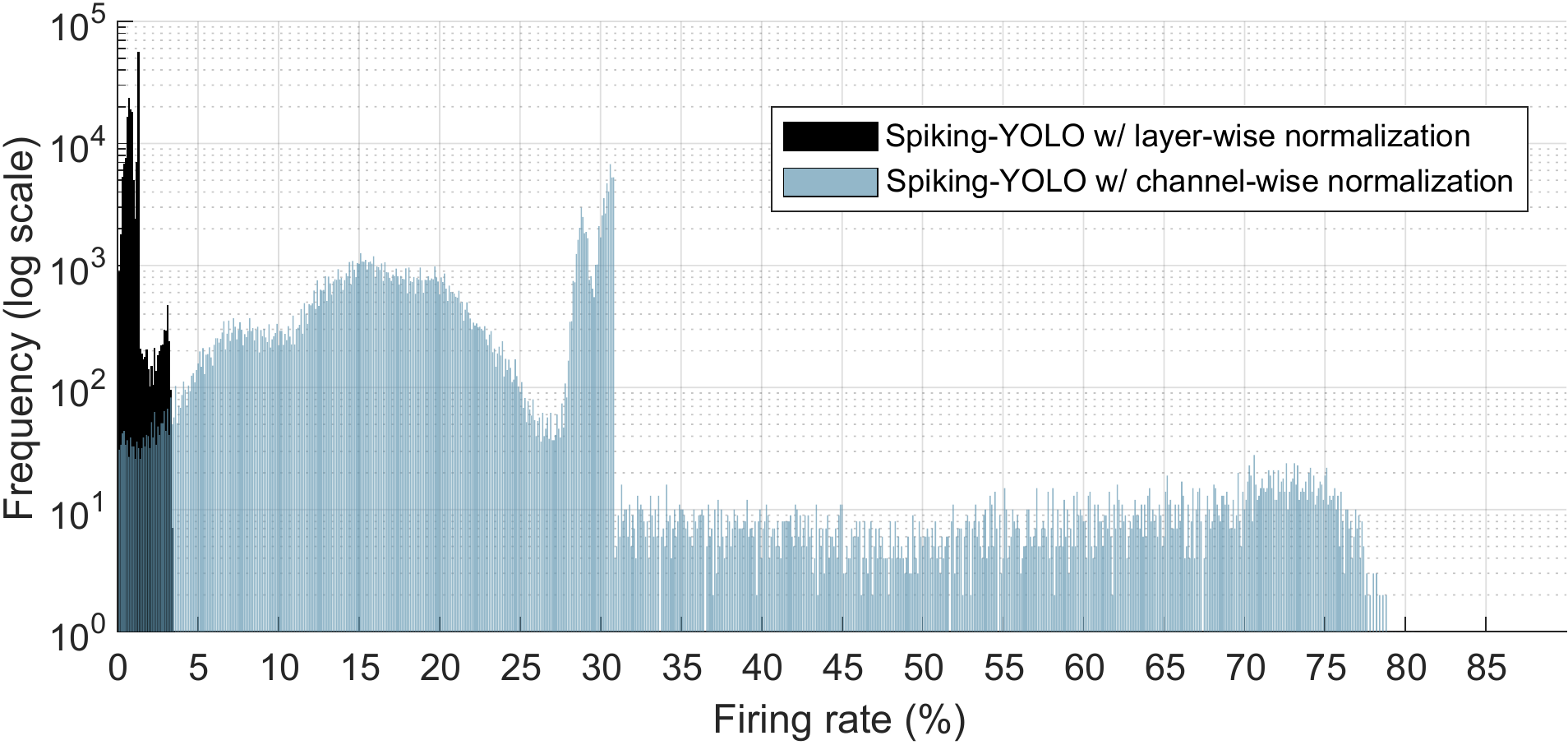}
    \caption{Firing rate distribution for layer-norm and channel-norm on channel 2 of Conv1 layer (Tiny YOLO)}
    \label{fig:spike_count_distribution}
    %\vspace{-0.5em}
\end{figure}

\textbf{Proposed normalization method } We propose a more fine-grained normalization method, called channel-wise normalization (abbreviated to channel-norm) to enable fast and efficient information transmission in deep SNNs. Our method normalizes the weights by the maximum possible activation (the 99.9th percentile) in a channel-wise manner instead of the conventional layer-wise manner. The proposed channel-norm can be expressed as

\begin{equation}
\label{eq:channel-wise norm}
\tilde{w}^{l}_{i,j} = w^{l}_{i,j} \frac{\lambda^{l-1}_{i}}{\lambda^{l}_{j}} \quad \textrm{and} \quad \tilde{b}^{l}_{j} = \frac{b^{l}_{j}}{\lambda^{l}_{j}}, 
\end{equation} where $i$ and $j$ are indices of channels. Weights $w$ in a layer $l$ are normalized (same effect as normalizing the output activation) by maximum activation $\lambda^{l}_{j}$ in each channel. As mentioned before, the maximum activation is calculated from the training dataset. In the following layer, the normalized activations must be multiplied by $\lambda^{l-1}_{i}$ to obtain the original activation prior to the normalization. The detailed method is depicted in Algorithm 1 and Figure~\ref{fig:channel-norm}. 

Normalizing the activations in the channel-wise manner eliminates extremely small activations (i.e., under-activation), which had small activation values prior to the normalization. In other words, neurons are normalized to obtain a higher yet proper firing rate, which leads to accurate information transmission in a short period of time.

\begin{figure}[t]
    \centering
    \includegraphics[width=\columnwidth]{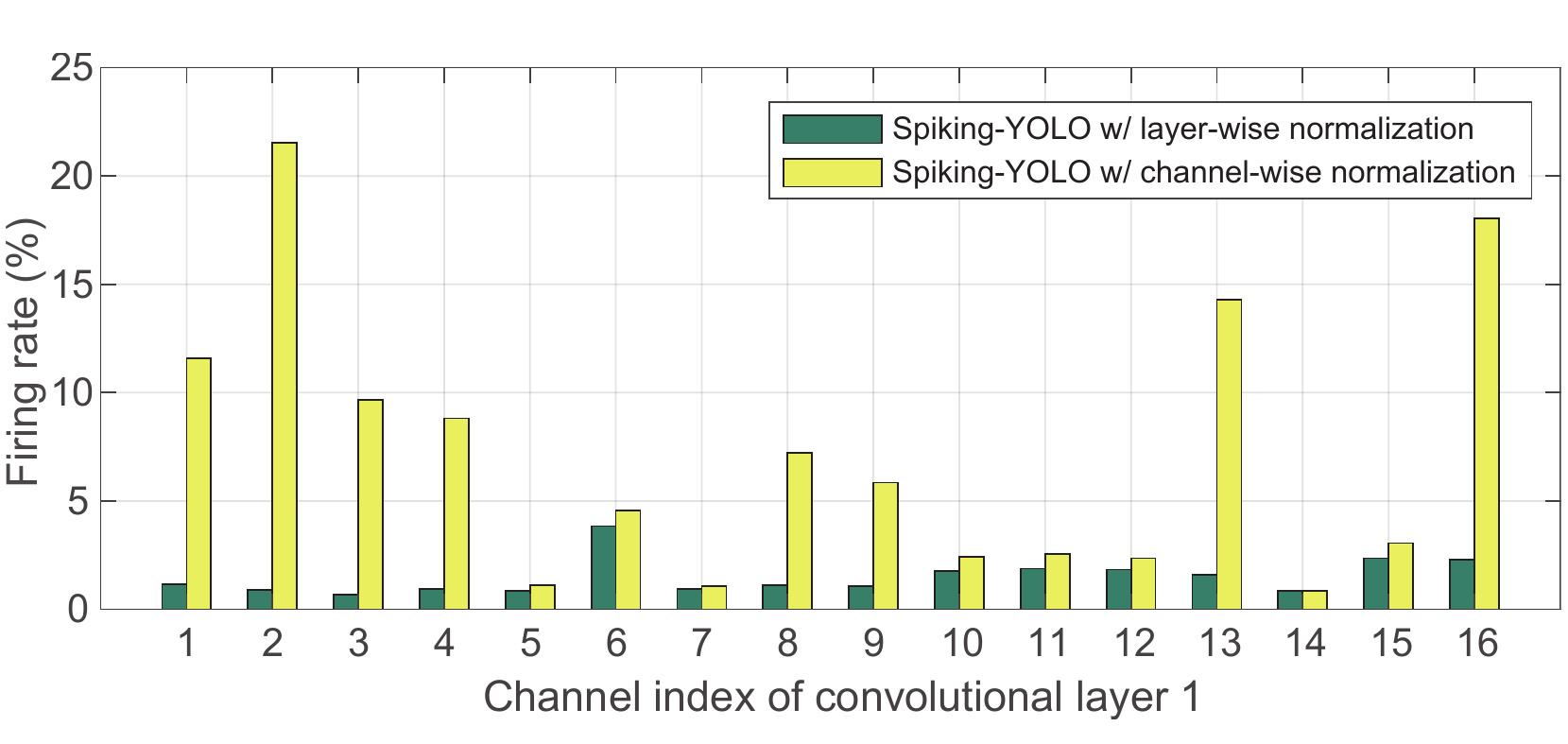} \caption{Firing rate of 16 channels in Conv1 layer for layer-norm and channel-norm of Tiny YOLO}
%    \vspace{-0.5em}
    \label{fig:avg_firing_rate}
\end{figure}

\begin{figure}[h]
    \vspace{-0.5em}
    \centering
    \includegraphics[width=.9\columnwidth]{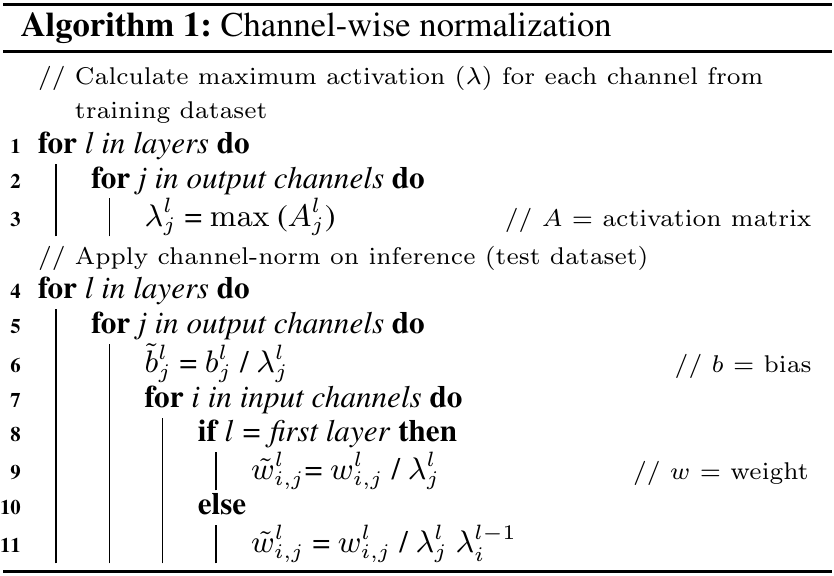} 
    \vspace{-0.5em}
\end{figure}

\begin{figure}[t]
    \center
    \includegraphics[width=.96\columnwidth]{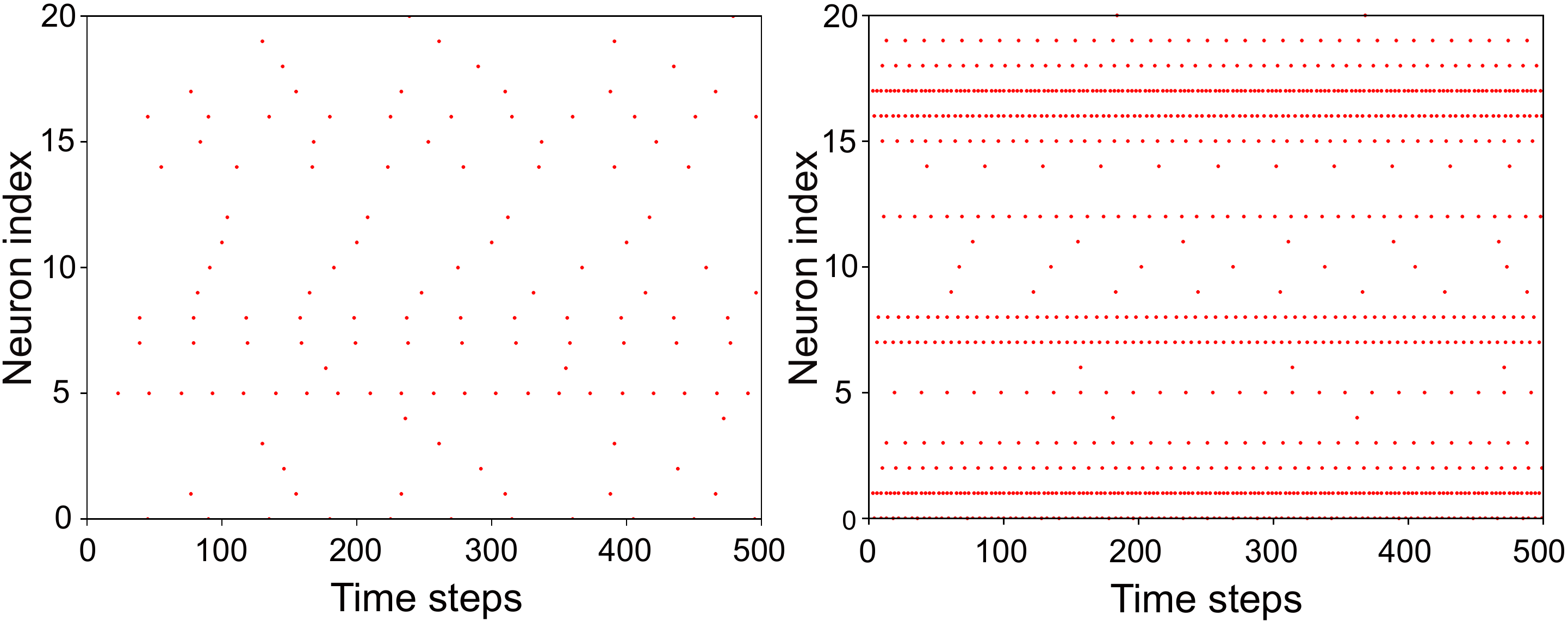}
    \caption{Raster plot of 20 sampled neurons' spike activity; layer-norm (left) vs. channel-norm (right)}
	\label{fig:raster_plot}
	%\vspace{-0.5em}
\end{figure}

\textbf{Analysis of the improved firing rate } In Figure~\ref{fig:spike_count_distribution}, the x- and y-axes indicate the firing rate and the number of neurons that produce a specific firing rate on a log scale, respectively. For channel-norm, numerous neurons generated a firing rate of up to 80\%. In layer-norm, however, most of the neurons generated a firing rate in the range between 0\% and 3.5\%. This is a clear indication that channel-norm eliminates extremely small activations and that more neurons are producing a higher yet proper firing rate. In addition, Figure~\ref{fig:avg_firing_rate} presents the firing rate in each channel in the convolutional layer 1. Evidently, channel-norm produces a much higher firing rate in majority of the channels. Particularly in channel 2, channel-norm produces a firing rate that is 20 times higher than that of layer-norm. Furthermore, Figure~\ref{fig:raster_plot} presents a raster plot of the spike activity from 20 sampled neurons. It can be seen that numerous neurons are firing more regularly when channel-norm is applied.

Our detailed analysis verifies that the fine-grained channel-norm normalizes activations better, preventing insufficient activation that leads to a low firing rate. In other words, extremely small activations are normalized properly such that neurons can transmit information accurately in a short period of time. These small activations may not be significant and have little impact on the final output of the network in simple applications such as image classification; however, they are critical in regression problems and significantly affect the model's accuracy. Thus, channel-norm is a viable solution for solving more advanced machine learning problems in deep SNNs.

\subsection{Signed neuron featuring imbalanced threshold}

\textbf{Limitation of leaky-ReLU implementation in SNNs} ReLU, one of the most commonly used activation functions, retains solely positive input values and discards all negative values; $f(x) = x $ when $x \geq 0$, otherwise $f(x) = 0 $. Unlike ReLU, leaky-ReLU contains negative values with a leakage term, slope of $\alpha$, which is typically set to 0.01; $f(x) = x $ when $x \geq 0$, otherwise $f(x) = \alpha x$~\cite{xu2015empirical}. 

Most previous DNN-to-SNN conversion methods have focused on converting integrate-and-fire neurons to ReLU, while completely neglecting the leakage term in the negative region of the activation function. Note that negative activations account for over 51\% in Tiny YOLO. To extend the activation function bound to the negative region in SNNs, \cite{rueckauer2017conversion} added a second $V_\mathrm{th}$ term ($-1$). Their method successfully 
converted BinaryNet~\cite{hubara2016binarized} to SNNs, where the BinaryNet activations were constrained to $+1$ or $-1$ on CIFAR-10. 

Currently, various DNNs use leaky-ReLU as an activation function, yet an accurate and efficient method of implementing leaky-ReLU in an SNN domain has not been proposed. Leaky-ReLU can be implemented in SNNs by simply multiplying negative activations by the slope ${\alpha}$ in addition to a second $V_\mathrm{th}$ term ($-1$). However, this is not biologically plausible (the spike is a discrete signal) and can be a formidable challenge when employed on neuromorphic chips. For instance, additional hardware would be required for the floating-point multiplication of the slope ${\alpha}$.

\begin{figure}[t]
    \centering
    \includegraphics[width=.95\columnwidth]{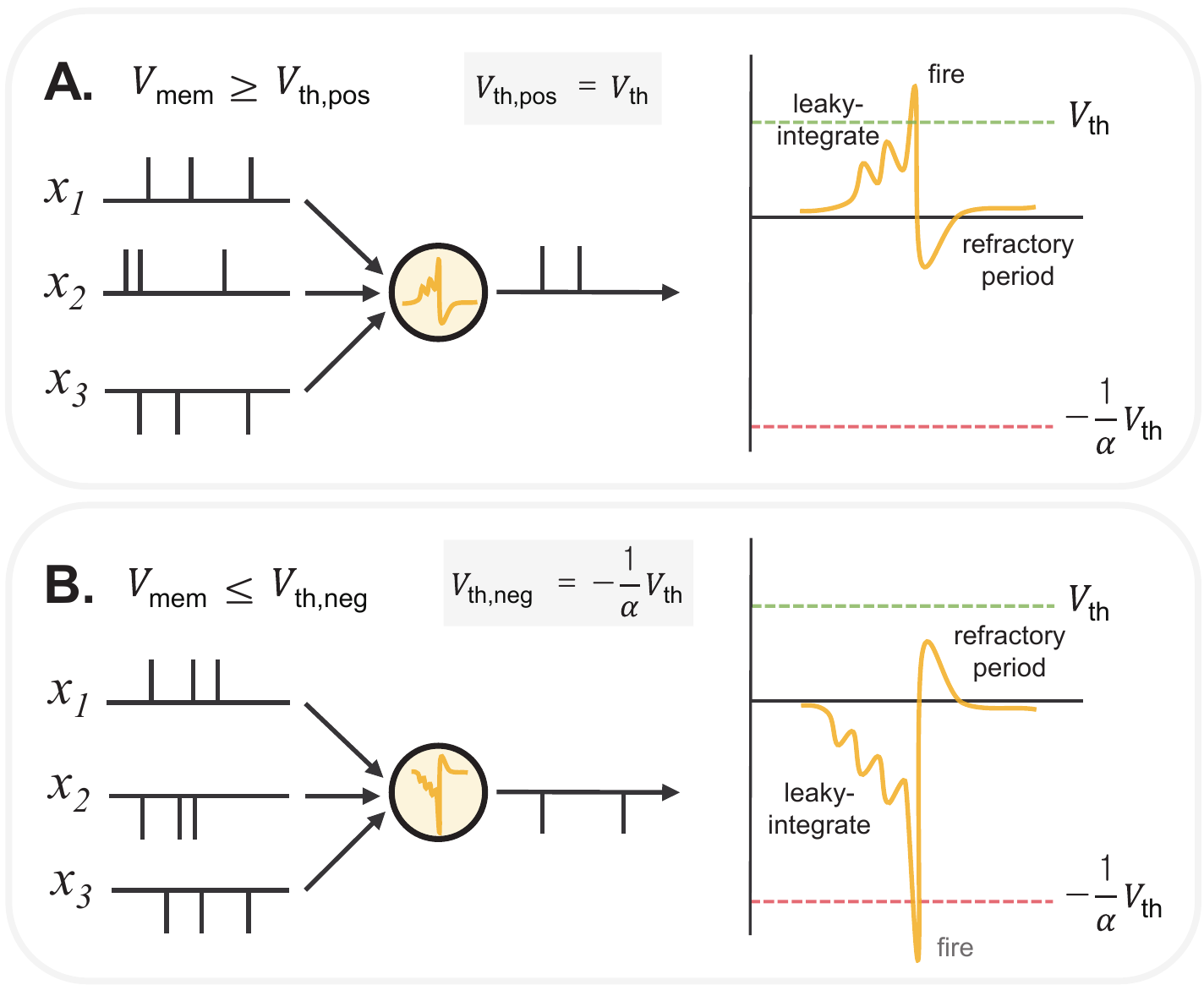} \caption{Overview of proposed signed neuron featuring imbalanced threshold; two possible cases for a spiking neuron}
    \label{fig:signed_neuron}
%    \vspace{-0.5em}
\end{figure}

\textbf{The notion of imbalanced threshold } We herein introduce a signed neuron featuring imbalanced threshold (hereinafter abbreviated as IBT) that can not only interpret both positive and negative activations, but also accurately and efficiently compensate for the leakage term in the negative regions of leaky-ReLU. The proposed method also retains the discrete characteristics of the spikes by introducing a different threshold voltage for the negative region, $V_\mathrm{th,neg}$. The second threshold voltage $V_\mathrm{th,neg}$ is equal to the $V_\mathrm{th}$ divided by the negative of the slope, $-\alpha$, and $V_\mathrm{th,pos}$ is equal to $V_\mathrm{th}$ as before. This would replicate the leakage term (slope $\alpha$) in the negative region of leaky-ReLU. The underlying dynamics of signed neuron with IBT are represented by 
\begin{equation}
\label{eq:leaky-relu-SNN}
 \mathrm{fire}(V_{\mathrm{mem}}) =
  \begin{cases}
    1 &  \text{if} \;\; V_{\mathrm{mem}} \geq V_{\mathrm{th,pos}} (V_{\mathrm{th}}) \\
    -1 &  \text{if} \;\; V_{\mathrm{mem}} \leq V_{\mathrm{th,neg}} (-\frac{1}{\alpha} V_{\mathrm{th}}) \\
    0 &  \textrm{otherwise,\;\:no firing.}
  \end{cases}
\end{equation} 
As shown in Figure~\ref{fig:signed_neuron}, if the slope $\alpha = 0.1$ then the threshold voltage responsible for a positive activation $V_\mathrm{th,pos}$ is $1V$, and that for a negative activation, $V_\mathrm{th,neg}$, is $-10V$; therefore, $V_\mathrm{mem}$ must be integrated ten times more to generate a spike for the negative activations in leaky-ReLU.

It is noteworthy that a signed neuron also enables implementation of excitatory and inhibitory neurons, which is more biologically plausible \cite{dehghani2016dynamic,wilson1972excitatory}. Using signed neurons with IBT, leaky-ReLU can be implemented accurately in SNNs and can directly be mapped to the current neuromorphic architecture with minimum overhead. Moreover, the proposed method will create more opportunities for converting various DNN models to SNNs in a wide range of applications. 

\begin{figure*} [t]
    \centering
    \includegraphics[width=.95\columnwidth]{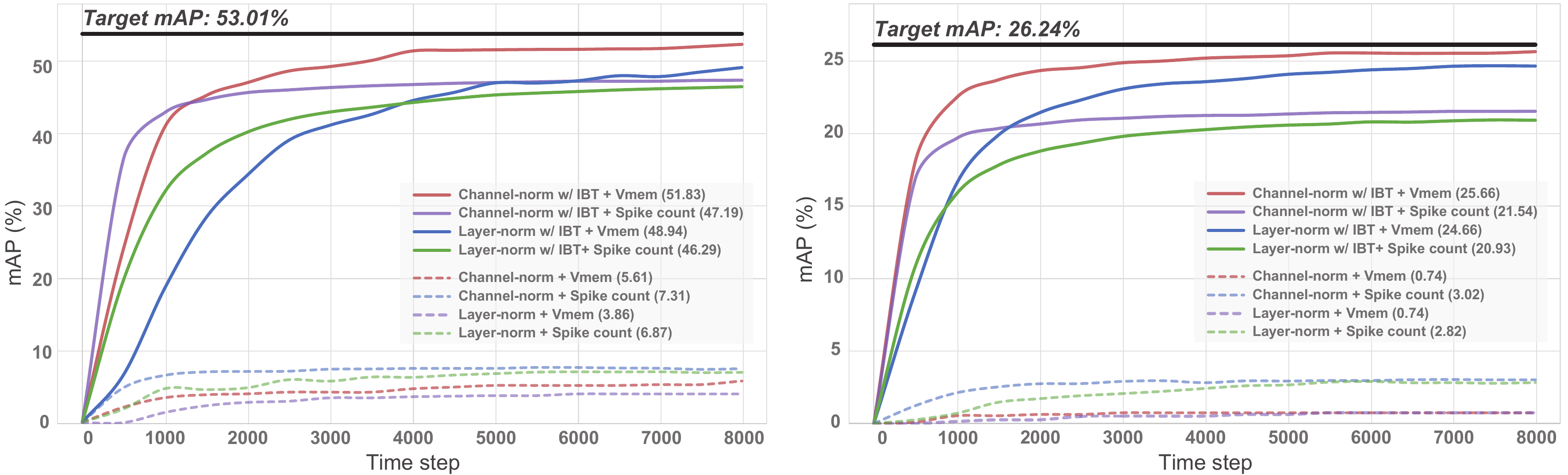}
    \caption{Experimental results of Spiking-YOLO on PASCAL VOC (left) and MS COCO (right) for various configurations (normalization methods + signed neuron w/ IBT + decoding scheme); maximum mAP is in parentheses.}
    \label{fig:yolo_snn_results}
    %\vspace{-0.5em}
\end{figure*}

\section{Evaluation}
\subsection{Experimental setup}
As the first step towards object detection in deep SNNs, we used a real-time object detection model, Tiny YOLO, which is a simpler but efficient version of YOLO. We implemented max-pooling and batch-normalization in SNNs according to \cite{rueckauer2017conversion}. Tiny YOLO is tested on non-trivial datasets, PASCAL VOC and MS COCO. Our simulation is based on the TensorFlow Eager and we conducted all experiments on NVIDIA Tesla V100 GPUs.

\subsection{Experimental results}
\textbf{Spiking-YOLO detection results} To verify and analyze the functionalities of the proposed methods, we investigated the effects of the presence or absence of channel-norm and signed neuron with IBT. As depicted in Figure~\ref{fig:yolo_snn_results}, when both channel-norm and signed neuron with IBT are applied, Spiking-YOLO achieves a remarkable performance of 51.83\% and 25.66\% on VOC PASCAL and MS COCO, respectively. The target mAP of Tiny YOLO is 53.01\% (PASCAL VOC) and 26.24\% (MS COCO). In fact, channel-norm outperforms layer-norm in detecting objects by a large margin, especially on PASCAL VOC (53.01\% vs. 48.94\%), and converges faster. For instance, to reach the maximum mAP of layer-norm (48.94), channel norm only requires approximately 3,500 time steps (2.3x faster). Similar results are observed in MS COCO where channel-norm converges even faster than the layer-norm (4x faster). Please refer to Table~\ref{table:yolo_results} for more detailed results.

\renewcommand\theadalign{bc}
\renewcommand\cellgape{\Gape[3pt]}
\floatsetup[table]{capposition=top}
\begin{table}[h]
    \vspace{-0.5em}
    \resizebox{\columnwidth}{!}{
    \begin{threeparttable}
        \begin{tabular}{llcccc}
            \toprule
            \multirow{2}{*}{\makecell{Signed \\ neuron}} & \multirow{2}{*}{\makecell{Norm. \\ method}} &
            \multicolumn{2}{c}{PASCAL VOC (53.01)\tnote{a}} & \multicolumn{2}{c}{MS COCO (26.24)\tnote{a}}\\ 
            \cmidrule(r){3-4}
            \cmidrule(r){5-6}
            && $V_{\mathrm{mem}}$ & Spike count & $V_{\mathrm{mem}}$ & Spike count \\
            \midrule
            \multirow{2}{*}{w/out IBT} & Layer & 3.86 & 6.87 &  0.74 & 2.82\\
            & Channel & 5.61 & 7.31 & 0.74 & 3.02\\
            \midrule
            \multirow{2}{*}{w/ IBT} & Layer & 48.94 & 46.29 &  24.66 & 20.93\\
            & Channel & \textbf{51.83} & 47.19 & \textbf{25.66} & 21.54\\
            \bottomrule
        \end{tabular}
        \begin{tablenotes}
            \item[a] {Target mAP in parentheses}
        \end{tablenotes}
    \end{threeparttable}
    }
    \caption{Experiment results for Spiking-YOLO (mAP \%)}
    \label{table:yolo_results}
\end{table}

Notably, without the proposed methods, the model failed to detect objects, reporting 6.87\% and 2.82\% for VOC PASCAL and MS COCO, respectively. When channel-norm is applied, the model still struggles to detect objects, reporting approximately 7.31\% and 3.02\% at the best. This is a great indication that signed neuron with IBT accurately implements the leakage term in leaky-ReLU. Thus, the rest of the experiments were conducted using signed neuron with IBT as the default. 

For further analysis, we performed additional experiments on two different output decoding schemes: one based on accumulated $V_\mathrm{mem}$, and another based on spike count. The quotient from $V_\mathrm{mem}$ / $V_\mathrm{th}$ indicates the spike count, and the remainder is rounded off. This remainder will eventually become an error and lost information. Therefore, the $V_\mathrm{mem}$-based output decoding scheme is more precise for interpreting spike trains; Figure~\ref{fig:yolo_snn_results} verifies this assertion. The $V_\mathrm{mem}$-based output decoding scheme outperforms the spike-count-based scheme and converges faster in channel-norm.

\begin{figure*} [t]
    \centering
    \includegraphics[width=.95\columnwidth]{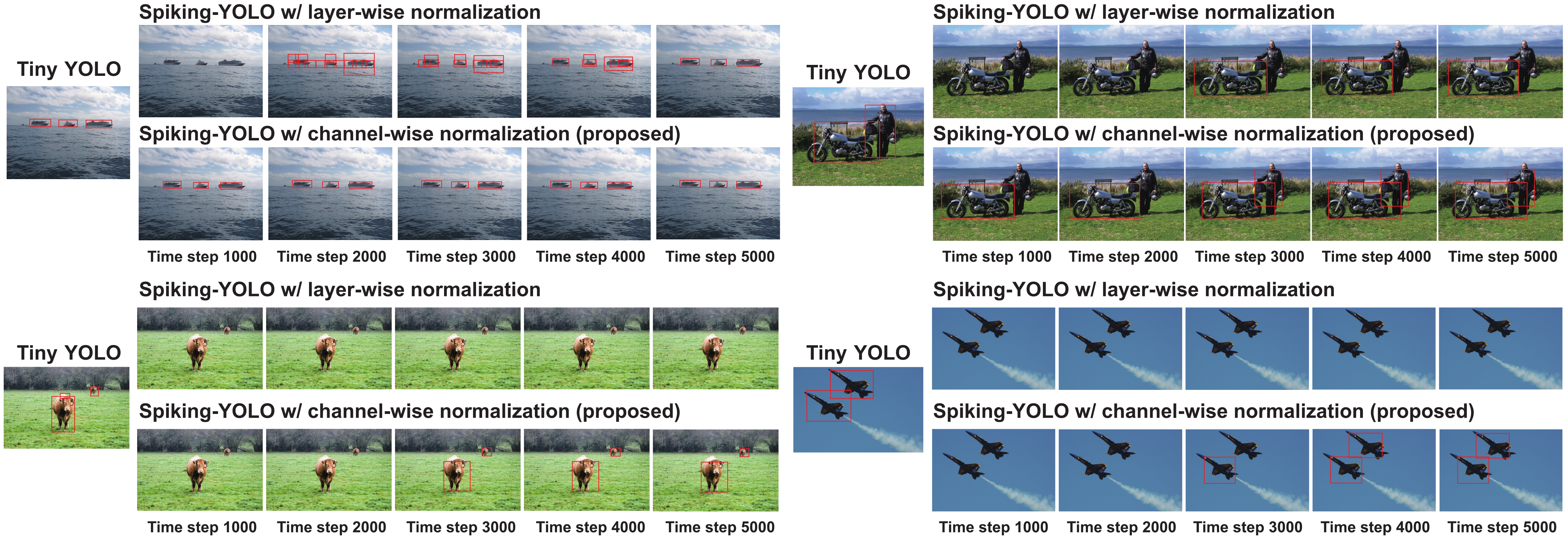}
    \caption{Object detection results (Tiny YOLO vs. Spiking-YOLO with layer-norm vs. Spiking-YOLO with channel-norm)}
    \label{fig:bounding_boxes_3}
\end{figure*}

Figure~\ref{fig:bounding_boxes_3} illustrates the efficacy of Spiking-YOLO in detecting objects as the time step increases. For each example, the far-left image (Tiny YOLO) shows the ground truth label that Spiking-YOLO attempts to replicate. In the top-left example (three ships), after only 1000 time steps, Spiking-YOLO with channel-norm successfully detects all three objects. Meanwhile, Spiking YOLO with layer-norm failed to detect any objects. After 2000 time steps, it starts to draw bounding boxes around the objects, but there are multiple bounding boxes drawn over a single object, and their sizes are all inaccurate. The detection performance improves as the time steps increase but is still unsatisfactory; 5000 time steps are required to reach the detection performance of the proposed channel-norm. This remarkable performance of Spiking-YOLO is also shown in the other examples in Figure~\ref{fig:bounding_boxes_3}. 
The proposed channel-norm shows a clear advantage in detecting multiple and microscopic objects accurately in a shorter period of time. 
%Please refer to the supplementary material for more object detection results.

\textbf{Spiking-YOLO energy efficiency} To investigate energy efficiency of Spiking-YOLO, we considered two different approaches: a) computing operations of Spiking-YOLO and Tiny YOLO in digital signal processing b) Spiking-YOLO on neuromorphic chips vs. Tiny YOLO on GPUs. 

\begin{figure} [h]
    \centering
    \includegraphics[width=.95\columnwidth]{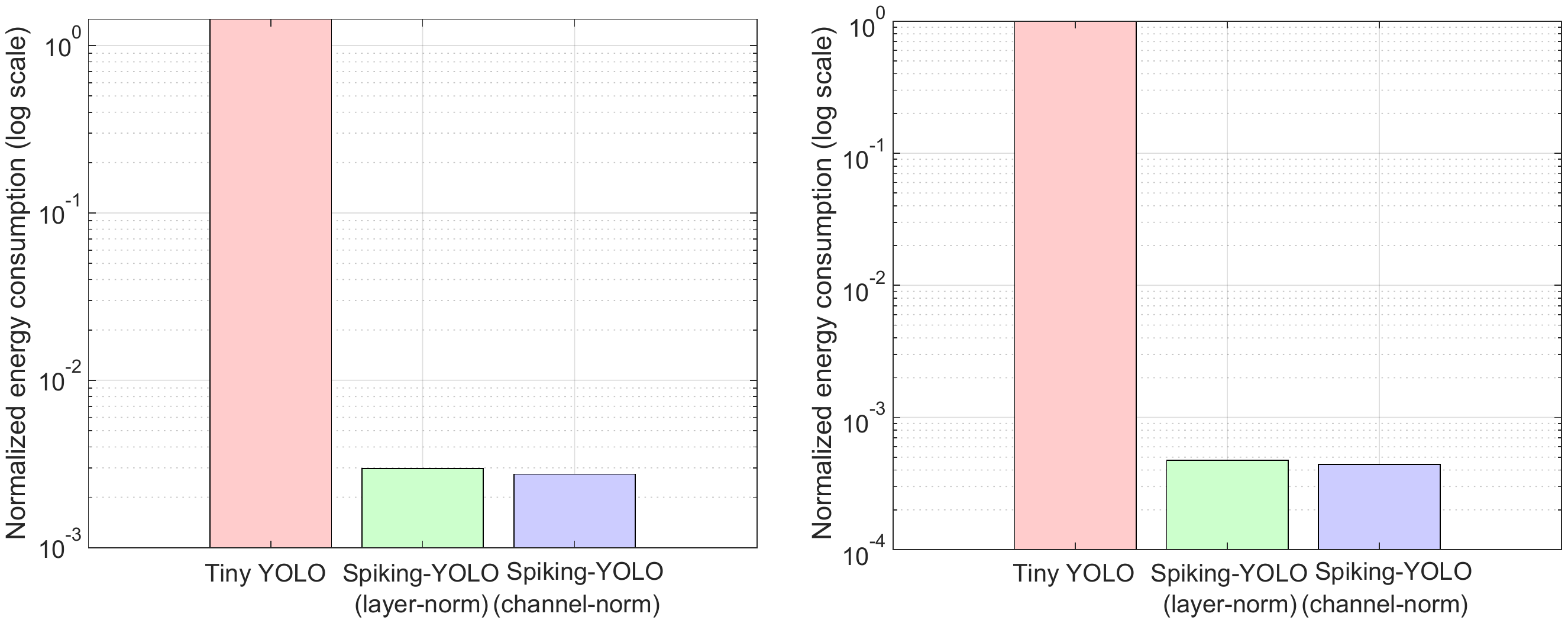}
\caption{Energy comparison of Tiny YOLO and Spiking-YOLO for MAC and AC operations; 32-bit FL (left) and 32-bit INT (right)}
	\label{fig:energy_consumption1}
	%\vspace{-0.5em}
\end{figure}

Firstly, most operations in DNNs occur in convolutional layers where the multiply-accumulate (MAC) operations are primarily responsible during execution. SNNs, however, perform accumulate (AC) operations because spike events are binary operations whose input is integrated (or accumulated) into a membrane potential only when spikes are received. For a fair comparison, we focused solely on the computational power (MAC and AC) used to execute object detection on a single image. 
According to ~\cite{horowitz20141}, a 32-bit floating-point (FL) MAC operation consumes 4.6 pJ (0.9 + 3.7 pJ) and 0.9 pJ for an AC operation. A 32-bit integer (INT) MAC operation consumes 3.2 pJ (0.1 + 3.1 pJ) and 0.1 pJ for an AC operation. Based on these measures, we calculated the energy consumption of Tiny YOLO and Spiking-YOLO by multiplying FLOPs (floating-point operations) and the energy consumption of MAC and AC operations calculated, as shown below. FLOPs of Tiny YOLO are reported on ~\cite{yolo_website}, and that for Spiking-YOLO are calculated during our simulation. Figure \ref{fig:energy_consumption1} shows that regardless of the normalization methods, Spiking-YOLO demonstrates exceptional energy efficiency, over 2,000 times better than Tiny-YOLO for 32-bit FL and INT operations.

Secondly, SNNs on neuromorphic chips offer excellent energy efficiency, which is an important and desirable aspect of neural networks~\cite{pfeiffer2018deep}. We compare the energy consumption of Tiny YOLO and Spiking-YOLO when each ran on the latest GPU (Titan V100) and neuromorphic chip (TrueNorth), respectively. The power and GFLOPS (Giga floating-point operation per second) of Titan V100 were obtained from \cite{nvidia2017v100}, and GFLOPS/W for TrueNorth is reported on \cite{merolla2014million}. We define one time step as equal to 1ms (1 kHz synchronization signal in \cite{merolla2014million}). 

Based on our calculations shown in Table \ref{table:energy_table_2}, Spiking-YOLO consumes approximately 280 times less energy than Tiny YOLO when ran on TrueNorth. As mentioned in the experimental results, the proposed channel-norm converges much faster than layer-norm; therefore, the energy consumption of Spiking-YOLO with channel-norm is approximately four times less than that with layer-norm as they have similar power consumption. Note that contemporary GPUs are far more advanced computing technology, and the TrueNorth chip was first introduced in 2014. As neuromorphic chips continue to develop and have better performance, we can expect even higher energy and computational efficiency. 

\begin{table}[h]
    \centering
    \resizebox{\columnwidth}{!}
    {\begin{tabular}{ccccc|c}
        \toprule
        \multicolumn{6}{c}{\textbf{Tiny YOLO}}\\
        \midrule
        & Power {\small (W)} & GFLOPS & FLOPs &  & Energy {\small (J)}\\
        \midrule
        & 250 & 14,000 &  6.97E+09 &  & 0.12 \\
        \midrule
        \multicolumn{6}{c}{\textbf{Spiking-YOLO}}\\

        \midrule
        \makecell{Norm. \\ methods} & \makecell{{\small GFLOPS} \\ {\small / W}} & FLOPs & Power {\small (W)} & \makecell{Time \\ steps} & Energy {\small (J)}\\
        \midrule
        Layer & 400 & 5.28E+07 & 1.320E-04 & 8,000 & 1.06E-03\\
        Channel & 400  & 4.90E+07 & 1.225E-04 & \textbf{3,500} & \textbf{4.29E-04} \\ 
        \bottomrule
    \end{tabular}}
    \caption{Energy comparison of Tiny YOLO (GPU) and Spiking-YOLO (neuromorphic chips)}
    \label{table:energy_table_2}
\end{table}

\section{Conclusion}
In this paper, we have presented Spiking-YOLO, the first SNN model that successfully performs object detection by achieving comparable results to those of the original DNNs on non-trivial datasets, PASCAL VOC and MS COCO. In doing so, we proposed two novel methods. We believe that this study represents the first step towards solving more advanced machine learning problems in deep SNNs. 

\section{Acknowledgements}
This work was supported in part by the National Research Foundation of Korea (NRF) grant funded by the Korean government (Ministry of Science and ICT) [2016M3A7B4911115, 2018R1A2B3001628], the Brain Korea 21 Plus Project in 2019, Samsung Electronics (DS and Foundry), and AIR Lab (AI Research Lab) in Hyundai Motor Company through HMC-SNU AI Consortium Fund.

\newpage
\bibliographystyle{aaai20}
\bibliography{aaai20.bib}

\end{document}